\documentclass{article}

\usepackage[final]{nips_2016}

\usepackage[utf8]{inputenc} 
\usepackage[T1]{fontenc} 
\usepackage{url} 
\usepackage{booktabs} 
\usepackage{amsfonts} 
\usepackage{nicefrac} 
\usepackage{microtype} 

\usepackage{times}

\usepackage{graphicx} 
\usepackage{subfig}

\usepackage[ruled]{algorithm}
\usepackage{algorithmic}

\usepackage{amssymb}
\usepackage{epstopdf}
\usepackage{epsfig}
\usepackage{amsmath}

\usepackage{caption}
\usepackage{verbatimbox}
\usepackage{tabularx}

\newtheorem{theorem}{Theorem}[section]

\newtheorem{definition}[theorem]{Definition}

\usepackage{hyperref}

\title{A Communication-Efficient Parallel Algorithm for Decision Tree}

\author{Qi Meng$^{1,}$\thanks{Denotes equal contribution. This work was done when the first author was visiting Microsoft Research Asia.} , Guolin Ke$^{2,*}$, Taifeng Wang$^2$, Wei Chen$^2$, Qiwei Ye$^2$,\\
	\textbf{Zhi-Ming Ma$^3$, Tie-Yan Liu$^2$}\\
	$^1$Peking University\quad$^2$Microsoft Research\\$^3$Chinese Academy of Mathematics and Systems Science	\\
	$^1$qimeng13@pku.edu.cn;
	$^2$\{Guolin.Ke, taifengw, wche, qiwye, tie-yan.liu\}@microsoft.com;\\
	$^3$mazm@amt.ac.cn
}

\begin{document}

\maketitle

\begin{abstract}
Decision tree (and its extensions such as Gradient Boosting Decision Trees and Random Forest) is a widely used machine learning algorithm, due to its practical effectiveness and model interpretability. With the emergence of big data, there is an increasing need to parallelize the training process of decision tree. However, most existing attempts along this line suffer from high communication costs. In this paper, we propose a new algorithm, called \emph{Parallel Voting Decision Tree (PV-Tree)}, to tackle this challenge. After partitioning the training data onto a number of (e.g., $M$) machines, this algorithm performs both local voting and global voting in each iteration. For local voting, the top-$k$ attributes are selected from each machine according to its local data. Then, globally top-$2k$ attributes are determined by a majority voting among these local candidates. Finally, the full-grained histograms of the globally top-$2k$ attributes are collected from local machines in order to identify the best (most informative) attribute and its split point. PV-Tree can achieve a very low communication cost (independent of the total number of attributes) and thus can scale out very well. Furthermore, theoretical analysis shows that this algorithm can learn a near optimal decision tree, since it can find the best attribute with a large probability. Our experiments on real-world datasets show that PV-Tree significantly outperforms the existing parallel decision tree algorithms in the trade-off between accuracy and efficiency.
\end{abstract}

\section{Introduction}
Decision tree [\cite{quinlan1986induction}] is a widely used machine learning algorithm, since it is practically effective and the rules it learns are simple and interpretable. Based on decision tree, people have developed other algorithms such as Random Forest (RF) [\cite{breiman2001random}] and Gradient Boosting Decision Trees (GBDT) [\cite{friedman2001greedy}], which have demonstrated very promising performances in various learning tasks [\cite{burges2010ranknet}].

In recent years, with the emergence of very big training data (which cannot be held in one single machine), there has been an increasing need of parallelizing the training process of decision tree. To this end, there have been two major categories of attempts: \footnote{There is another category of works that parallelize the tasks of sub-tree training once a node is split [\cite{pearson1993coarse}], which require the training data to be moved from machine to machine for many times and are thus inefficient. Moreover, there are also some other works accelerating decision tree construction by using pre-sorting [\cite{mehta1996sliq}, \cite{shafer1996sprint} \cite{joshi1998scalparc}] and binning [\cite{ranka1998clouds}, \cite{gehrke1999boat}, \cite{jin2003communication}], or employing a shared-memory-processors approach [\cite{kufrin1997decision} \cite{agrawal2001parallel}]. However, they are out of our scope.}.

\emph{Attribute-parallel}: Training data are vertically partitioned according to the attributes and allocated to different machines, and then in each iteration, the machines work on non-overlapping sets of attributes in parallel in order to find the best attribute and its split point (suppose this best attribute locates at the $i$-th machine) [\cite{shafer1996sprint}, \cite{joshi1998scalparc}, \cite{svore2011large}]. This process is communicationally very efficient. However, after that, the re-partition of the data on other machines than the $i$-th machine will induce very high communication costs (proportional to the number of data samples). This is because those machines have no information about the best attribute at all, and in order to fulfill the re-partitioning, they must retrieve the partition information of every data sample from the $i$-th machine. Furthermore, as each worker still has full sample set, the partition process is not parallelized, which slows down the algorithm.

\emph{Data-parallel}: Training data are horizontally partitioned according to the samples and allocated to different machines. Then the machines communicate with each other the local histograms of all attributes (according to their own data samples) in order to obtain the global attribute distributions and identify the best attribute and split point [\cite{kufrin1997decision}, \cite{panda2009planet}]. It is clear that the corresponding communication cost is very high and proportional to the total number of attributes and histogram size. To reduce the cost, in [\cite{ben2010streaming}, \cite{tyree2011parallel}, \cite{jin2003communication}], it was proposed to exchange quantized histograms between machines when estimating the global attribute distributions. However, this does not really solve the problem – the communication cost is still proportional to the total number of attributes, not to mentioned that the quantization may hurt the accuracy.

In this paper, we proposed a new data-parallel algorithm for decision tree, called \emph{Parallel Voting Decision Tree (PV-Tree)}, which can achieve much better balance between communication efficiency and accuracy. The key difference between conventional data-parallel decision tree algorithm and PV-Tree lies in that the former only trusts the globally aggregated histogram information, while the latter leverages the local statistical information contained in each machine through a two-stage voting process, thus can significantly reduce the communication cost. Specifically, PV-Tree contains the following steps in each iteration. 1) \emph{Local voting}. On each machine, we select the top-$k$ attributes based on its local data according to the informativeness scores (e.g., risk reduction for regression, and information gain for classification). 2) \emph{Global voting}. We determine global top-$2k$ attributes by a majority voting among the local candidates selected in the previous step. That is, we rank the attributes according to the number of local machines who select them, and choose the top $2k$ attributes from the ranked list. 3) \emph{Best attribute identification}. We collect the full-grained histograms of the globally top-$2k$ attributes from local machines in order to compute their global distributions. Then we identify the best attribute and its split point according to the informativeness scores calculated from the global distributions.

It is easy to see that PV-Tree algorithm has a very low communication cost. It does not need to communicate the information of all attributes, instead, it only communicates indices of the locally top-$k$ attributes per machine and the histograms of the globally top-$2k$ attributes. In other words, its communication cost is independent of the total number of attributes. This makes PV-Tree highly scalable. On the other hand, it can be proven that PV-Tree can find the best attribute with a large probability, and the probability will approach $1$ regardless of $k$ when the training data become sufficiently large. In contrast, the data-parallel algorithm based on quantized histogram could fail in finding the best attribute, since the bias introduced by histogram quantization cannot be reduced to zero even if the training data are sufficiently large.

We have conducted experiments on real-world datasets to evaluate the performance of PV-Tree. The experimental results show that PV-Tree has consistently higher accuracy and training speed than all the baselines we implemented. We further conducted experiments to evaluate the performance of PV-Tree in different settings (e.g., with different numbers of machines, different values of $k$). The experimental results are in accordance with our theoretical analysis.

\section{Decision Tree}

Suppose the training data set $D_n=\{(x_{i,j},y_i); i=1,\cdots,n, j=1,\cdots,d\}$ are independently sampled from $ \prod_{j=1}^d \mathcal{X}_j\times \mathcal{Y}$ according to $(\prod_{j=1}^d P_{X_j}) P_{Y|X}$. The goal is to learn a regression or classification model $f\in\mathcal{F}:\prod_{j=1}^d\mathcal{X}_j\to\mathcal{Y}$ by minimizing loss functions on the training data, which hopefully could achieve accurate prediction for the unseen test data.

Decision tree [\cite{quinlan1986induction,safavian1991survey}] is a widely used model for both regression [\cite{breiman1984classification}] and classification [\cite{safavian1991survey}]. A typical decision tree algorithm is described in Alg \ref{alg:BulidTree}. As can be seen, the tree growth procedure is recursive, and the nodes will not stop growing until they reach the \textit{stopping criteria}. There are two important functions in the algorithm: \textit{FindBestSplit} returns the best split point \textit{\{attribute, threshold\}} of a node,

\begin{minipage}[t]{0.6\textwidth}
\vspace{-10pt}

 and \textit{Split} splits the training data according to the best split point. The details of \textit{FindBestSplit} is given in Alg \ref{alg:FindBestSplit}: first histograms of the attributes are constructed (for continuous attributes, one usually converts their numerical values to finite bins for ease of compuation) by going over all training data on the current node; then all \textit{bins} (split points) are traversed from left to right, and \textit{leftSum} and \textit{rightSum} are used to accumulate sum of left and right parts of the split point respectively. When selecting the best split point, an informativeness measure is adopted. The widely used informative measures are \emph{information gain} and \emph{variance gain} for classification and regression, respectively.

\end{minipage}%
\hspace{3mm}
\begin{minipage}[t]{0.36\textwidth}
\vspace{-20pt}

\begin{algorithm}[H]
\caption{BulidTree}
\label{alg:BulidTree}
\begin{algorithmic}\small
\STATE {\bfseries Input:} Node N, Dateset D
\IF{StoppingCirteria(D)}
\STATE N.output = Prediction(D)
\ELSE
\STATE bestSplit = FindBestSplit(D)
\STATE (DL, DR) = Split(D, N, bestSplit)
\STATE BuildTree(N.leftChild, DL)
\STATE BuildTree(N.rightChild, DR)
\ENDIF
\end{algorithmic}
\end{algorithm}
\end{minipage}%

\begin{definition}\label{def1}[\cite{friedman2001elements},\cite{quinlan1986induction}]
	In classification, the information gain (IG) for attribute $X_j\in[w_1,w_2]$ at node $O$, is defined as the entropy reduction of the output $Y$ after splitting node $O$ by attribute $X_j$ at $w$, i.e.,
	{\small\begin{align*}
		IG_j(w;O)&=\mathcal{H}_j-(\mathcal{H}_j^l(w)+\mathcal{H}_j^r(w))\\
		&=P(w_1\leq X_j\leq w_2)H(Y|w_1\leq X_j\leq w_2)-P(w_1\leq X_j<w)H(Y|w_1\leq X_j<w)\\
		&\quad-P(w\leq X_j\leq w_2)H(Y|w\leq X_j\leq w_2),
		\end{align*}}
	where $H(\cdot|\cdot)$ denotes the conditional entropy.
	
	In regression, the variance gain (VG) for attribute $X_j\in[w_1,w_2]$ at node $O$, is defined as variance reduction of the output $Y$ after splitting node $O$ by attribute $X_j$ at $w$, i.e.,
	{\small\begin{align*}
		VG_j(w;O)&=\sigma_j-(\sigma_j^l(w)+\sigma_j^r(w))\\
		&=P(w_1\leq X_j\leq w_2)Var[Y|w_1\leq X_j\leq w_2]-P(w_1\leq X_j<w)Var[Y|w_1\leq X_j<w]\\
		&\quad-P(w_2\geq X_j\geq w)Var[Y|w_2\geq X_j\geq w],
		\end{align*}}
	where $Var[\cdot|\cdot]$ denotes the conditional variance.
\end{definition}

\section{PV-Tree}

In this section, we describe our proposed PV-Tree algorithm for parallel decision tree learning, which has a very low communication cost, and can achieve a good trade-off between communication efficiency and learning accuracy.

PV-Tree is a data-parallel algorithm, which also partitions the training data onto $M$ machines just like in [\cite{ben2010streaming},\cite{tyree2011parallel}]. However, its design principal is very different. In [\cite{ben2010streaming},\cite{tyree2011parallel}], one does not trust the local information about the attributes in each machine, and decides the best attribute and split point only based on the aggregated global histograms of the attributes. In contrast, in PV-Tree, we leverage the meaningful statistical information about the attributes contained in each local machine, and make decisions through a two-stage (local and then global) voting process. In this way, we can significantly reduce the communication cost since we do not need to communicate the histogram information of all the attributes across machines, instead, only the histograms of those attributes that survive in the voting process.

The flow of PV-tree algorithm is very similar to the standard decision tree, except function \textit{FindBestSplit}. So we only give the new implementation of this function in Alg \ref{alg:PV-Tree-FindBestSplit}, which contains following three steps:

\emph{Local Voting}: We select the top-$k$ attributes for each machine based on its local data set (according to the informativeness scores, e.g., information gain for classification and variance reduction for regression), and then exchange indices of the selected attributes among machines. Please note that the communication cost for this step is very low, because only the indices for a small number of (i.e., $k\times M$) attributes need to be communicated.

\emph{Global Voting}: We determine the globally top-$2k$ attributes by a majority voting among all locally selected attributes in the previous step. That is, we rank the attributes according to the number of local machines who select them, and choose the top-$2k$ attributes from the ranked list. It can be proven that when the local data are big enough to be statistically representative, there is a very high probability that the top-$2k$ attributes obtained by this majority voting will contain the globally best attribute. Please note that this step does not induce any communication cost.

\emph{Best Attribute Identification}: We collect full-grained histograms of the globally top-$2k$ attributes from local machines in order to compute their global distributions. Then we identify the best attribute and its split point according to the informativeness scores calculated from the global distributions. Please note that the communication cost for this step is also low, because we only need to communicate the histograms of $2k$ pre-selected attributes (but not all attributes).\footnote{As indicated by our theoretical analysis and empirical study (see the next sections), a very small $k$ already leads to good performance in PV-Tree algorithm.} As a result, PV-Tree algorithm can scale very well since its communication cost is independent of both the total number of attributes and the total number of samples in the dataset.

In next section, we will provide theoretical analysis on accuracy guarantee of PV-Tree algorithm.

\begin{minipage}[t]{0.50\textwidth}
\vspace{-20pt}
\begin{algorithm}[H]
\caption{FindBestSplit}
\label{alg:FindBestSplit}
\begin{algorithmic}\small
\STATE {\bfseries Input:} DataSet D
\FORALL{X in D.Attribute}
\STATE $\triangleright$ \textit{Construct Histogram}
\STATE H = new Histogram()
\FORALL{x in X}
\STATE H.binAt(x.bin).Put(x.label)
\ENDFOR
\STATE $\triangleright$ \textit{Find Best Split}
\STATE leftSum = new HistogramSum()
\FORALL{bin in H}
\STATE leftSum = leftSum + H.binAt(bin)
\STATE rightSum = H.AllSum - leftSum
\STATE split.gain = CalSplitGain(leftSum, rightSum)
\STATE bestSplit = ChoiceBetterOne(split,bestSplit)
\ENDFOR
\ENDFOR
\STATE \textbf{return} bestSplit
\end{algorithmic}
\end{algorithm}
\end{minipage}%
\hspace{1mm}
\begin{minipage}[t]{0.48\textwidth}
\vspace{-20pt}
\begin{algorithm}[H]
\caption{PV-Tree\_FindBestSplit}
\label{alg:PV-Tree-FindBestSplit}
\begin{algorithmic}\small
\STATE {\bfseries Input:} Dataset D
\STATE localHistograms = ConstructHistograms(D)
\STATE $\triangleright$ \textit{Local Voting}
\STATE splits = []
\FORALL {H in localHistograms}
\STATE splits.Push(H.FindBestSplit())
\ENDFOR
\STATE localTop = splits.TopKByGain(K)
\STATE $\triangleright$ \textit{Gather all candidates}
\STATE allCandidates = AllGather(localTop)
\STATE $\triangleright$ \textit{Global Voting}
\STATE globalTop = allCandidates.TopKByMajority(2*K)
\STATE $\triangleright$ \textit{Merge global histograms}
\STATE globalHistograms = Gather(globalTop, localHistograms)
\STATE bestSplit = globalHistograms.FindBestSplit()
\STATE \textbf{return} bestSplit
\end{algorithmic}
\end{algorithm}
\end{minipage}

\section{Theoretical Analysis}

In this section, we conduct theoretical analysis on proposed PV-Tree algorithm. Specifically, we prove that, PV-Tree can select the best (most informative) attribute in a large probability, for both classification and regression.
In order to better present the theorem, we firstly introduce some notations\footnote{Since all analysis are for one arbitrarily fixed node $O$, we omit the notation $O$ here.} In classification, we denote {\small$IG_j=\max_w{IG_j(w)}$}, and rank $\{IG_j;j\in[d]\}$ from large to small as {\small$\{IG_{(1)},...,IG_{(d)}\}$}. We call the attribute $j_{(1)}$ the most informative attribute. Then, we denote {\small$l_{(j)}(k)=\frac{|IG_{(1)}-IG_{(j)}|}{2}, \forall j\geq k+1$} to indicate the distance between the largest and the $k$-th largest IG. In regression, $l_{(j)}(k)$ is defined in the same way, except replacing IG with VG.
\begin{theorem}
	\label{regression_thm}
	Suppose we have $M$ local machines, and each one has $n$ training data. PV-Tree at an arbitrary tree node with local voting size $k$ and global majority voting size $2k$ will select the most informative attribute with a probability at least
	{\small\begin{eqnarray}\label{regressionbound}
		\sum_{m=[M/2+1]}^MC_M^m\left(1-\left(\sum_{j=k+1}^d\delta_{(j)}(n,k)\right)\right)^m\left(\sum_{j=k+1}^d\delta_{(j)}(n,k)\right)^{M-m}\nonumber,
		\end{eqnarray}}
	where {\small$\delta_{(j)}(n,k) =\alpha_{(j)}(n)+4e^{-c_{(j)}n\left(l_{(j)}(k)\right)^2}$} with {\small$\lim_{n\rightarrow\infty}{\alpha_{(j)}(n)}=0$} and $c_{(j)}$ is constant.
\end{theorem}

Due to space restrictions, we briefly illustrate the proof idea here and leave detailed proof to supplementary materials. Our proof contains two parts. (1) For local voting, we find a sufficient condition to guarantee a similar rank of attributes ordered by information gain computed based on local data and full data. Then, we derive a lower bound of probability to make the sufficient condition holds by using concentration inequalities. (2) For global voting, we select top-$2k$ attributes. It’s easy to proof that we can select the most informative attribute if only no less than {\small$[M/2+1]$} of all machines select it.\footnote{In fact, the global voting size can be $\beta k$ with $\beta>1$. Then the sufficient condition becomes that no less than $[M/\beta+1]$ of all machines select the most informative attribute.} Therefore, we can calculate the probability in the theorem using binomial distribution.

Regarding Theorem \ref{regression_thm}, we have following discussions on factors that impact the lower bound for probability of selecting the best attribute.

1.\emph{Size of local training data $n$:} Since $\delta_{(j)}(n,k)$ decreased with $n$,
with more and more local training data, the lower bound will increase. That means, if we have sufficiently large data, PV-Tree will select the best attribute with almost probability $1$.

2. \emph{Input dimension $d$:} It is clear that for fixed local voting size $k$ and global voting size $2k$, with $d$ increasing, the lower bound is decreasing. Consider the case that the number of attributes become $100$ times larger. Then the terms in the summation (from $\sum_{j=k+1}^d$ to $\sum_{j=k+1}^{100d}$) is roughly $100$ times larger for a relatively small $k$. But there must be many attributes away from attribute $(1)$ and $l_{(j)}(k)$ is a large number which results in a small $\delta_{(j)}(n,k)$. Thus we can say that the bound in the theorem is not sensitive with $d$.

3. \emph{Number of machines $M$:} We assume the whole training data size $N$ is fixed and the local data size $n=\frac{N}{M}$. Then on one hand, as $M$ increases, $n$ decreases, and therefore the lower bound will decrease due to larger $\delta_j(n,k)$. On the other hand, because function $\sum_{m=[M/2+1]}^{M}C_M^mp^m(1-p)^{M-m}$ will approach $1$ as $M$ increases when $p>0.5$ [\cite{zhou2012ensemble}], the lower bound will increase. In other words, the number of machines $M$ has dual effect on the lower bound: with more machines, local data size becomes smaller which reduces the accuracy of local voting, however, it also leads to more copies of local votes and thus increase the reliability of global voting. Therefore, in terms of accuracy, there should be an optimal number of machines given a fixed-size training data.\footnote{Please note that using more machines will reduce local computing time, thus the optimal value of machine number may be larger in terms of speed-up.}

4. \emph{Local/Global voting size $k/2k$:} Local/Global voting size $k/2k$ influence $l_{(j)}(k)$ and the terms in the summation in the lower bound . As $k$ increases, $l_{(j)}(k)$ increases and the terms in the summation decreases, and the lower bound increases. But increasing $k$ will bring more communication and calculating time. Therefore, we should better select a moderate $k$. For some distributions, especially for the distributions over high-dimensional space, $l_{(j)}(k)$ is less sensitive to $k$, then we can choose a relatively smaller $k$ to save communication time.

As a comparison, we also prove a theorem for the data-parallel algorithm based on quantized histogram as follows (please refer to the supplementary material for its proof). The theorem basically tells us that the bias introduced by histogram quantization cannot be reduced to zero even if the training data are sufficiently large, and as a result the corresponding algorithm could fail in finding the best attribute.\footnote{The theorem for regression holds in the same way, with replacing IG with VG.} This could be the critical weakness of this algorithm in big data scenario.

\begin{theorem}
	We denote quantized histogram with $b$ bins of the underlying distribution $P$ as $P^b$, that of the empirical distribution $P_n$ as $P_n^b$, the information gain of $X_j$ calculated under the distribution $P^b$ and $P_n^b$ as $IG_j^b$ and $IG_{n,j}^b$ respectively, and $f_j(b)\triangleq|IG_j-IG^b_{j}|$. Then, for $\epsilon \leq \min_{j=1,\cdots,d} f_j(b)$, with probability at least $\delta_{j}(n, f_j(b)-\epsilon))$, we have $|IG_{n,j}^b -IG_j|>\epsilon$.
\end{theorem}

\section{Experiments}

In this section, we report the experimental comparisons between PV-Tree and baseline algorithms. We used two data sets, one for learning to rank (LTR) and the other for ad click prediction (CTR)\footnote{We use private data in LTR experiments and data of KDD Cup 2012 track 2 in CTR experiments.} (see Table \ref{useddata} for details). For LTR, we extracted about 1200 numerical attributes per data sample, and used NDCG [\cite{burges2010ranknet}] as the evaluation measure. For CTR, we extracted about 800 numerical attributes [\cite{jahrer2012ensemble}], and used AUC as the evaluation measure.

\begin{table}[H]\small
	\setlength{\tabcolsep}{4.2pt}
	\renewcommand{\arraystretch}{1}
\begin{minipage}[t]{0.48\textwidth}
\vspace{0pt}
\begin{small}
\caption{Datasets}
\centering
\begin{tabular}{ccccc}
	\hline
	
	\hline
Task&\#Train&\#Test&\#Attribute&Source \\
&&&&\\\hline
LTR&11M&1M&1200&Private \\
CTR&235M&31M&800&KDD Cup \\ \hline

\hline
\label{useddata}
\end{tabular}
\end{small}
\end{minipage}
\begin{minipage}[t]{0.48\textwidth}
\vspace{0pt}
\begin{small}
\caption{Convergence time (seconds)}
\centering
\begin{tabular}{ccccc}
	\hline
	
	\hline
Task&Sequential&Data-&Attribute-&PV-Tree\\
&&Parallel&Parallel &\\ \hline
LTR&28690&32260&14660&5825\\
CTR&154112&9209&26928&5349\\
    \hline

    \hline
\label{convergespeed}
\end{tabular}
\end{small}
\end{minipage}
\end{table}
\vspace{-20px}
According to recent industrial practices, a single decision tree might not be strong enough to learn an effective model for complicated tasks like ranking and click prediction. Therefore, people usually use decision tree based boosting algorithms (e.g., GBDT) to perform tasks. In this paper, we also use GBDT as a platform to examine the efficiency and effectiveness of decision tree parallelization. That is, we used PV-Tree or other baseline algorithms to parallelize the decision tree construction process in each iteration of GBDT, and compare their performance. Our experimental environment is a cluster of servers (each with 12 CPU cores and 32 GB RAM) inter-connected with 1 Gbps Ethernet. For the experiments on LTR, we used 8 machines for parallel training; and for the experiments on CTR, we used 32 machines since the dataset is much larger.

\subsection{Comparison with Other Parallel Decision Trees}

For comparison with PV-Tree, we have implemented an attribute-parallel algorithm, in which a binary vector is used to indicate the split information and exchanged across machines. In addition, we implemented a data-parallel algorithm according to [\cite{ben2010streaming,tyree2011parallel}], which can communicate both full-grained histograms and quantized histograms. All parallel algorithms and sequential(single machine) version are compared together.

The experimental results can be found in Figure \ref{ranking_overall} and \ref{ctr_overall}. From these figures, we have the following observations:

For LTR, since the number of data samples is relatively small, the communication of the split information about the samples does not take too much time. As a result, the attribute-parallel algorithm appears to be efficient. Since most attributes take numerical values in this dataset, the full-grained histogram has quite a lot of bins. Therefore, the data-parallel algorithm which communicates full-grained histogram is quite slow, even slower than the sequential algorithm. When reducing the bins in the histogram to 10\%, the data-parallel algorithm becomes much more efficient, however, its convergence point is not good (consistent with our theory – the bias in quantized histograms leads to accuracy drop).

For CTR, attribute-parallel algorithm becomes very slow since the number of data samples is very large. In contrast, many attributes in CTR take binary or discrete values, which make the full-grained histogram have limited number of bins. As a result, the data-parallel algorithm with full-grain histogram is faster than the sequential algorithm. The data-parallel algorithm with quantized histograms is even faster, however, its convergence point is once again not very good.

PV-Tree reaches the best point achieved by sequential algorithm within the shortest time in both LTR and CTR task. For a more quantitative comparison on efficiency, we list the time for each algorithm (8 machines for LTR and 32 machines for CTR) to reach the convergent accuracy of the sequential algorithm in Table \ref{convergespeed}. From the table, we can see that, for LTR, it costed PV-Tree 5825 seconds, while it costed the data-parallel algorithm (with full-grained histogram\footnote{The data-parallel algorithm with 10\% bins could not achieve the same accuracy with the sequential algorithm and thus we did not put it in the table.}) and attribute-parallel algorithm 32260 and 14660 seconds respectively. As compared with the sequential algorithm (which took 28690 seconds to converge), PV-Tree achieves 4.9x speed up on 8 machines. For CTR, it costed PV-Tree 5349 seconds, while it costed the data-parallel algorithm (with full-grained histogram) and attribute-parallel algorithm 9209 and 26928 seconds respectively. As compared with the sequential algorithm (which took 154112 seconds to converge), PV-Tree achieves 28.8x speed up on 32 machines.

We also conducted independent experiments to get a clear comparison of communication cost for different parallel algorithms given some typical big data workload setting. The result is listed in Table \ref{commcost}. We find the cost of attribute-parallel algorithm is relative to the size of training data $N$, and the cost of data-parallel algorithm is relative to the number of attributes $d$. In contrast, the cost of PV-Tree is constant.

\begin{table}[t]\small
		\setlength{\tabcolsep}{4.4pt}
		\renewcommand{\arraystretch}{1}
\begin{minipage}[t]{0.45\textwidth}
\vspace{0pt}
\begin{small}
\caption{Comparison of communication \\ cost, train one tree with depth=6.}
\centering
\begin{tabular}{lccc} \hline
	
	\hline
Data size &Attribute &Data &PV-Tree\\
&Palallel &Parallel& k=15 \\ \hline
N=1B,&750MB&424MB&10MB \\
d=1200&&& \\ \hline
N=100M,&75MB&424MB&10MB \\
d=1200&&& \\ \hline
N=1B,&750MB&70MB&10MB \\
d=200&&& \\ \hline
N=100M,&75MB&70MB&10MB \\
d=200&&& \\ \hline

\hline
\label{commcost}
\end{tabular}
\end{small}
\end{minipage}
\begin{minipage}[t]{0.52\textwidth}
\vspace{0pt}
\caption{Convergence time and accuracy w.r.t. global voting parameter $k$ for PV-Tree.}
\centering
\begin{small}
\begin{tabular}{lccccc} \hline
	
	\hline
	&&&&&\\
&k=1&k=5&k=10&k=20&k=40\\
\hline
LTR&11256/&9906/&9065/&8323/&9529/\\
M=4&0.7905&0.7909&0.7909&0.7909&0.7909\\ \hline
LTR&8211/&8131/&8496/&10320/&12529/\\
M=16 &0.7882&0.7893&0.7897&0.7906&0.7909\\ \hline
CTR&9131/&9947/&9912/&10309/&10877/\\
M=16&0.7535&0.7538&0.7538&0.7538&0.7538\\ \hline
CTR&1806/&1745/&2077/&2133/&2564/\\
M=128&0.7533&0.7536&0.7537&0.7537&0.7538\\ \hline

\hline
\label{globalvote}
\end{tabular}
\end{small}
\end{minipage}
\end{table}

\begin{small}
\begin{figure}[tb]
\centering
\vspace{-20pt}
\subfloat[LTR, 8 machines]
{
\includegraphics[width=0.48\columnwidth, height=1.3in]{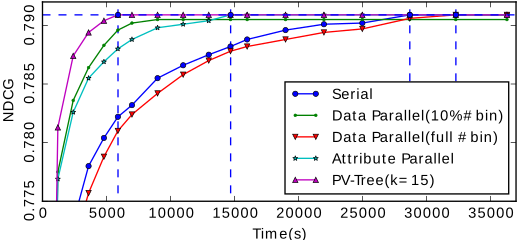}
\label{ranking_overall}
}
\subfloat[CTR, 32 machines]
{
\includegraphics[width=0.48\columnwidth, height=1.3in]{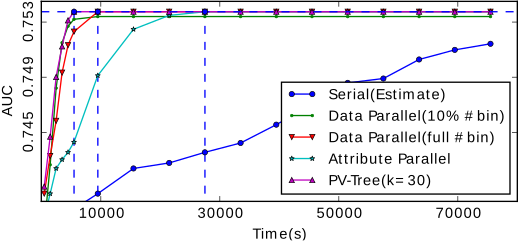}
\label{ctr_overall}
}
\caption{Performances of different algorithms}
\label{overall}
\vspace{-10pt}
\end{figure}
\end{small}
\subsection{Tradeoff between Speed-up and Accuracy in PV-Tree}
In the previous subsection, we have shown that PV-tree is more efficient than other algorithms. Here we make a deep dive into PV-tree to see how its key parameters affect the trade-off between efficiency and accuracy. According to Theorem \ref{regression_thm}, the following two parameters are critical to PV-Tree: the number of machines $M$ and the size of voting $k$.

\subsubsection {On Different Numbers of Machines}
When more machines join the distributed training process, the data throughput will grow larger but the amortized training data on each machine will get smaller. When the data size on each machine becomes too small, there will be no guarantee on the accuracy of the voting procedure, according to our theorem. So it is important to appropriately set the number of machines.

To gain more insights on this, we conducted some additional experiments, whose results are shown in Figure \ref{ranking_scaleoutmachinenumber} and \ref{ctr_scaleoutmachinenumber}. From these figures, we can see that for LTR, when the number of machines grows from 2 to 8, the training process is significantly accelerated. However, when the number goes up to 16, the convergence speed is even lower than that of using 8 machines. Similar results can be observed for CTR. These observations are consistent with our theoretical findings. Please note that PV-Tree is designed for the big data scenario. Only when the entire training data are huge (and thus distribution of the training data on each local machine can be similar to that of the entire training data), the full power of PV-Tree can be realized. Otherwise, we need to have a reasonable expectation on the speed-up, and should choose to use a smaller number of machines to parallelize the training.
\begin{small}
\begin{figure}[tb]
\centering
\vspace{0pt}
\subfloat[LTR]
{
\includegraphics[width=0.48\columnwidth, height=1.25in]{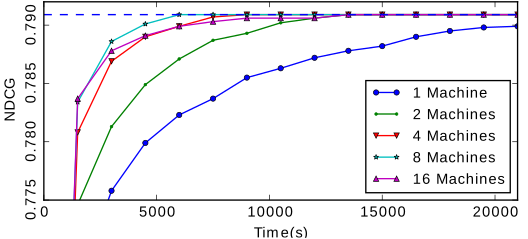}
\label{ranking_scaleoutmachinenumber}
}
\subfloat[CTR]
{
\includegraphics[width=0.48\columnwidth, height=1.25in]{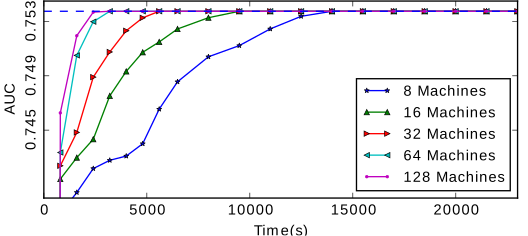}
\label{ctr_scaleoutmachinenumber}
}
\caption{PV-Tree on different numbers of machines}
\label{scaleoutmachinenumber}

\end{figure}
\end{small}

\subsubsection {On Different Sizes of Voting} \label{ExperimentsGlobalVote}

In PV-Tree, we have a parameter $k$, which controls the number of top attributes selected during local and global voting. Intuitively, larger $k$ will increase the probability of finding the globally best attribute from the local candidates, however, it also means higher communication cost. According to our theorem, the choice of $k$ should depend on the size of local training data. If the size of local training data is large, the locally best attributes will be similar to the globally best one. In this case, one can safely choose a small value of $k$. Otherwise, we should choose a relatively larger $k$. To gain more insights on this, we conducted some experiments, whose results are shown in Table \ref{globalvote}, where $M$ refers to the number of machines. From the table, we have the following observations. First, for both cases, in order to achieve good accuracy, one does not need to choose a large $k$. When $k \leq 40$, the accuracy has been very good. Second, we find that for the cases of using small number of machines, $k$ can be set to an even smaller value, e.g., $k=5$. This is because, given a fixed-size training data, when using fewer machines, the size of training data per machine will become larger and thus a smaller $k$ can already guarantee the approximation accuracy.

\subsection{Comparison with Other Parallel GBDT Algorithms}

While we mainly focus on how to parallelize the decision tree construction process inside GBDT in the previous subsections, one could also parallelize GBDT in other ways. For example, in [\cite{yu2001parallelizing,svore2011large}], each machine learns its own decision tree separately without communication. After that, these decision trees are aggregated by means of winner-takes-all or output ensemble. Although these works are not the focus of our paper, it is still interesting to compare with them.

For this purpose, we implemented both the algorithms proposed in [\cite{yu2001parallelizing}] and [\cite{svore2011large}]. For ease of reference, we denote them as \textit{Svore} and \textit{Yu} respectively. Their performances are shown in Figure \ref{ranking_vsboosting} and \ref{ctr_vsboosting}. From the figures, we can see that PV-Tree outperforms both \textit{Svore} and \textit{Yu}: although these two algorithms converge at a similar speed to PV-Tree, they have much worse converge points. According to our limited understanding, these two algorithms are lacking solid theoretical guarantee. Since the candidate decision trees are trained separately and independently without necessary information exchange, they may have non-negligible bias, which will lead to accuracy drop at the end. In contrast, we can clearly characterize the theoretical properties of PV-tree, and use it in an appropriate setting so as to avoid observable accuracy drop.

To sum up all the experiments, we can see that with appropriately-set parameters, PV-Tree can achieve a very good trade-off between efficiency and accuracy, and outperforms both other parallel decision tree algorithms designed specifically for GBDT parallelization.

\begin{small}
\begin{figure}[tb]
\centering
\vspace{-20pt}
\subfloat[LTR, 8 machines]
{
\includegraphics[width=0.48\columnwidth, height=1.25in]{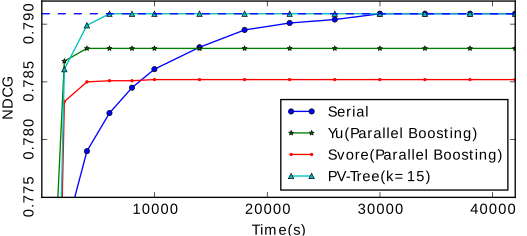}
\label{ranking_vsboosting}
}
\subfloat[CTR, 32 machines]
{
\includegraphics[width=0.48\columnwidth, height=1.25in]{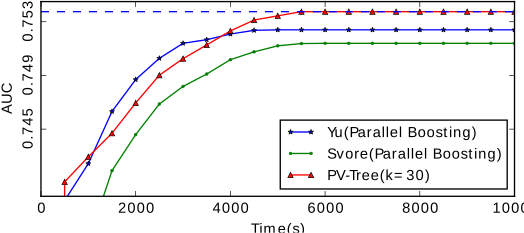}
\label{ctr_vsboosting}
}
\caption{Comparison with parallel boosting algorithms}
\label{vsboosting}
\vspace{-10pt}
\end{figure}
\end{small}

\section{Conclusions}

In this paper, we proposed a novel parallel algorithm for decision tree, called Parallel Voting Decision Tree (PV-Tree), which can achieve high accuracy at a very low communication cost. Experiments on both ranking and ad click prediction indicate that PV-Tree has its advantage over a number of baselines algorithms. As for future work, we plan to generalize the idea of PV-Tree to parallelize other machine learning algorithms. Furthermore, we will open-source PV-Tree algorithm to benefit more researchers and practitioners.



%
\small

\bibliographystyle{aaai}
\bibliography{pvtree} 

\begin{thebibliography}{}

\bibitem[\protect\citeauthoryear{Agrawal, Ho, and
  Zaki}{2001}]{agrawal2001parallel}
Agrawal, R.; Ho, C.-T.; and Zaki, M.~J.
\newblock 2001.
\newblock Parallel classification for data mining in a shared-memory
  multiprocessor system.
\newblock US Patent 6,230,151.

\bibitem[\protect\citeauthoryear{Banerjee, McKeague, and
  others}{2007}]{banerjee2007confidence}
Banerjee, M.; McKeague, I.~W.; et~al.
\newblock 2007.
\newblock Confidence sets for split points in decision trees.
\newblock {\em The Annals of Statistics} 35(2):543--574.

\bibitem[\protect\citeauthoryear{Ben-Haim and Tom-Tov}{2010}]{ben2010streaming}
Ben-Haim, Y., and Tom-Tov, E.
\newblock 2010.
\newblock A streaming parallel decision tree algorithm.
\newblock {\em The Journal of Machine Learning Research} 11:849--872.

\bibitem[\protect\citeauthoryear{Breiman \bgroup et al\mbox.\egroup
  }{1984}]{breiman1984classification}
Breiman, L.; Friedman, J.; Stone, C.~J.; and Olshen, R.~A.
\newblock 1984.
\newblock {\em Classification and regression trees}.
\newblock CRC press.

\bibitem[\protect\citeauthoryear{Breiman}{2001}]{breiman2001random}
Breiman, L.
\newblock 2001.
\newblock Random forests.
\newblock In {\em Machine learning}, volume~45,  5--32.
\newblock Springer.

\bibitem[\protect\citeauthoryear{Burges}{2010}]{burges2010ranknet}
Burges, C.~J.
\newblock 2010.
\newblock From ranknet to lambdarank to lambdamart: An overview.
\newblock In {\em Learning}, volume~11,  23--581.

\bibitem[\protect\citeauthoryear{Friedman, Hastie, and
  Tibshirani}{2001}]{friedman2001elements}
Friedman, J.; Hastie, T.; and Tibshirani, R.
\newblock 2001.
\newblock {\em The elements of statistical learning}, volume~1.
\newblock Springer series in statistics Springer, Berlin.

\bibitem[\protect\citeauthoryear{Friedman}{2001}]{friedman2001greedy}
Friedman, J.~H.
\newblock 2001.
\newblock Greedy function approximation: a gradient boosting machine.
\newblock In {\em Annals of statistics},  1189--1232.
\newblock JSTOR.

\bibitem[\protect\citeauthoryear{Gehrke \bgroup et al\mbox.\egroup
  }{1999}]{gehrke1999boat}
Gehrke, J.; Ganti, V.; Ramakrishnan, R.; and Loh, W.-Y.
\newblock 1999.
\newblock Boat—optimistic decision tree construction.
\newblock In {\em ACM SIGMOD Record}, volume~28,  169--180.
\newblock ACM.

\bibitem[\protect\citeauthoryear{Jahrer \bgroup et al\mbox.\egroup
  }{2012}]{jahrer2012ensemble}
Jahrer, M.; Toscher, A.; Lee, J.; Deng, J.; Zhang, H.; and Spoelstra, J.
\newblock 2012.
\newblock Ensemble of collaborative filtering and feature engineered models for
  click through rate prediction.
\newblock In {\em KDDCup Workshop}.

\bibitem[\protect\citeauthoryear{Jin and Agrawal}{2003}]{jin2003communication}
Jin, R., and Agrawal, G.
\newblock 2003.
\newblock Communication and memory efficient parallel decision tree
  construction.
\newblock In {\em SDM},  119--129.
\newblock SIAM.

\bibitem[\protect\citeauthoryear{Joshi, Karypis, and
  Kumar}{1998}]{joshi1998scalparc}
Joshi, M.~V.; Karypis, G.; and Kumar, V.
\newblock 1998.
\newblock Scalparc: A new scalable and efficient parallel classification
  algorithm for mining large datasets.
\newblock In {\em Parallel processing symposium, 1998. IPPS/SPDP 1998},
  573--579.
\newblock IEEE.

\bibitem[\protect\citeauthoryear{Kufrin}{1997}]{kufrin1997decision}
Kufrin, R.
\newblock 1997.
\newblock Decision trees on parallel processors.
\newblock In {\em Machine Intelligence and Pattern Recognition}, volume~20,
  279--306.
\newblock Elsevier.

\bibitem[\protect\citeauthoryear{Mehta, Agrawal, and
  Rissanen}{1996}]{mehta1996sliq}
Mehta, M.; Agrawal, R.; and Rissanen, J.
\newblock 1996.
\newblock Sliq: A fast scalable classifier for data mining.
\newblock In {\em Advances in Database Technology—EDBT'96}. Springer.
\newblock  18--32.

\bibitem[\protect\citeauthoryear{Panda \bgroup et al\mbox.\egroup
  }{2009}]{panda2009planet}
Panda, B.; Herbach, J.~S.; Basu, S.; and Bayardo, R.~J.
\newblock 2009.
\newblock Planet: massively parallel learning of tree ensembles with mapreduce.
\newblock In {\em Proceedings of the VLDB Endowment}, volume~2,  1426--1437.
\newblock VLDB Endowment.

\bibitem[\protect\citeauthoryear{Pearson}{1993}]{pearson1993coarse}
Pearson, R.~A.
\newblock 1993.
\newblock {\em A coarse grained parallel induction heuristic}.
\newblock University College, University of New South Wales, Department of
  Computer Science, Australian Defence Force Academy.

\bibitem[\protect\citeauthoryear{Quinlan}{1986}]{quinlan1986induction}
Quinlan, J.~R.
\newblock 1986.
\newblock Induction of decision trees.
\newblock In {\em Machine learning}, volume~1,  81--106.
\newblock Springer.

\bibitem[\protect\citeauthoryear{Ranka and Singh}{1998}]{ranka1998clouds}
Ranka, S., and Singh, V.
\newblock 1998.
\newblock Clouds: A decision tree classifier for large datasets.
\newblock In {\em Knowledge discovery and data mining},  2--8.

\bibitem[\protect\citeauthoryear{Safavian and
  Landgrebe}{1991}]{safavian1991survey}
Safavian, S.~R., and Landgrebe, D.
\newblock 1991.
\newblock A survey of decision tree classifier methodology.
\newblock {\em IEEE transactions on systems, man, and cybernetics}
  21(3):660--674.

\bibitem[\protect\citeauthoryear{Shafer, Agrawal, and
  Mehta}{1996}]{shafer1996sprint}
Shafer, J.; Agrawal, R.; and Mehta, M.
\newblock 1996.
\newblock Sprint: A scalable parallel classi er for data mining.
\newblock In {\em Proc. 1996 Int. Conf. Very Large Data Bases},  544--555.
\newblock Citeseer.

\bibitem[\protect\citeauthoryear{Svore and Burges}{2011}]{svore2011large}
Svore, K.~M., and Burges, C.
\newblock 2011.
\newblock Large-scale learning to rank using boosted decision trees.
\newblock {\em Scaling Up Machine Learning: Parallel and Distributed
  Approaches} 2.

\bibitem[\protect\citeauthoryear{Tyree \bgroup et al\mbox.\egroup
  }{2011}]{tyree2011parallel}
Tyree, S.; Weinberger, K.~Q.; Agrawal, K.; and Paykin, J.
\newblock 2011.
\newblock Parallel boosted regression trees for web search ranking.
\newblock In {\em Proceedings of the 20th international conference on World
  wide web},  387--396.
\newblock ACM.

\bibitem[\protect\citeauthoryear{Yu and Skillicorn}{2001}]{yu2001parallelizing}
Yu, C., and Skillicorn, D.
\newblock 2001.
\newblock Parallelizing boosting and bagging.
\newblock {\em Queen’s University, Kingston, Canada, Tech. Rep}.

\bibitem[\protect\citeauthoryear{Zhou}{2012}]{zhou2012ensemble}
Zhou, Z.-H.
\newblock 2012.
\newblock {\em Ensemble methods: foundations and algorithms}.
\newblock CRC Press.

\end{thebibliography}
\newpage
\section{Appendices}
	First of all, we review the definitions of information gain in classification and variance gain in regression.
	
	\textbf{Definition 2.1} [\cite{friedman2001elements},\cite{quinlan1986induction}]
		\textit{In classification, the information gain (IG) for attribute $X_j\in[w_1,w_2]$ at node $O$, is defined as the entropy reduction of the output $Y$ after splitting node $O$ by attribute $X_j$ at $w$, i.e.,}
		{\small\begin{align*}
			IG_j(w;O)&=\mathcal{H}_j-(\mathcal{H}_j^l(w)+\mathcal{H}_j^r(w))\\
			&=P(w_1\leq X_j\leq w_2)H(Y|w_1\leq X_j\leq w_2)-P(w_1\leq X_j<w)H(Y|w_1\leq X_j<w)\\
			&\quad-P(w\leq X_j\leq w_2)H(Y|w\leq X_j\leq w_2),
			\end{align*}}
	\textit{where $H(\cdot|\cdot)$ denotes the conditional entropy.}
		
		\textit{In regression, the variance gain (VG) for attribute $X_j\in[w_1,w_2]$ at node $O$, is defined as variance reduction of the output $Y$ after splitting node $O$ by attribute $X_j$ at $w$, i.e.,}
		{\small\begin{align*}
			VG_j(w;O)&=\sigma_j-(\sigma_j^l(w)+\sigma_j^r(w))\\
			&=P(w_1\leq X_j\leq w_2)Var[Y|w_1\leq X_j\leq w_2]-P(w_1\leq X_j<w)Var[Y|w_1\leq X_j<w]\\
			&\quad-P(w_2\geq X_j\geq w)Var[Y|w_2\geq X_j\geq w],
			\end{align*}}
		\textit{where $Var[\cdot|\cdot]$ denotes the conditional variance.}

	The conditional entropy $H(\cdot|\cdot)$ and the conditional variance $Var(\cdot|\cdot)$ are calculated according to the conditional distribution $P(\cdot|\cdot)$. For $K$ class classification, we assume $Y$ is a discrete random variable which takes value from the set $\{1,\cdots,K\}$ and we have
	\small{\begin{eqnarray}
		H(Y|w_1\leq X_j\leq w_2)&=&-\mathbb{E}_{(Y|w_1\leq X_j\leq w_2)}\log{p(Y|w_1\leq X_j\leq w_2)}\\
		&=&-\sum_{k=1}^Kp(Y=k|w_1\leq X_j\leq w_2)\log{p(Y=k|w_1\leq X_j\leq w_2)}.
		\end{eqnarray}}
	For regression, we assume that $Y$ is a continuous random variable and
	\small{\begin{eqnarray}
		Var(Y|w_1\leq X_j\leq w_2)&=&\mathbb{E}\left[(Y-\mathbb{E}[Y|w_1\leq X_j\leq w_2)]^2\big|w_1\leq X_j\leq w_2\right]\\
		&=&\int p(y|w_1\leq X_j\leq w_2)y^2dy-\left(\int p(y|w_1\leq X_j\leq w_2)ydy\right)^2.
		\end{eqnarray}}
	
	\subsection{Theorem 4.1 and its Proof for classification and regression}
	\textbf{Theorem 4.1:}
	\textit{In classification, suppose we have $M$ local machines, and each one has $n$ training data. PV-Tree at an arbitrary tree node with local voting size $k$ and global majority voting size $2k$ will select the most informative attribute with a probability at least
		{\small\begin{eqnarray}
			\sum_{m=[M/2+1]}^MC_M^m\left(1-\left(\sum_{j=k+1}^d\delta_{(j)}(n,k)\right)\right)^m\left(\sum_{j=k+1}^d\delta_{(j)}(n,k)\right)^{M-m}\nonumber,
			\end{eqnarray}}
		where {\small$\delta_{(j)}(n,k) =\alpha_{(j)}(n)+4e^{-c_{(j)}n\left(l_{(j)}(k)\right)^2}$} with {\small$\lim_{n\rightarrow\infty}{\alpha_{(j)}(n)}=0$} and $c_{(j)}$ is constant. }
	
	\vspace{5mm}
	\textbf{Proof for classification:}
	
	Firstly we introduce some notations. We use subscript $n$ to denote the corresponding empirical statistics, which is calculated based on the empirical distribution $\mathbb{P}_n$. Let $w_j^*=argmax_wIG_j(w)$ and $w_{n,j}^*=argmax_wIG_{n,j}(w)$. We denote $IG_{j}(w_j^*)$ as $IG_j$, which is the largest information gain for attribute $j$. We denote $IG_{n,j}(w_{n,j}^*)$ as $IG_{n,j}$, which is the largest empirical information gain for attribute $j$.
	As we defined in the main paper, we denote the index of attribute with the $j$-th largest information gain as $(j)$, and its corresponding information gain as $IG_{(j)}$, i.e.,
	{\small$$IG_{(1)}\geq\cdots\geq IG_{(j)}\geq\cdots\geq IG_{(d)}.$$}
	The corresponding empirical information gain for attribute $(j)$ denoted as $$IG_{n,(1)},...,IG_{n,(j)},...,IG_{n,(d)}.$$ Note that $IG_{n,(1)},...,IG_{n,(j)},...,IG_{n,(d)}$ may not be in an increasing order. Similarly, we denote the index of attribute with the $j$-th largest empirical information gain as $(j')$, and its corresponding empirical information gain as $IG_{n,(j')}$,i.e.,
	{\small$$IG_{n,(1')}\geq\cdots\geq IG_{n,(j')}\geq\cdots\geq IG_{n,(d')}.$$}
	Our proof idea is as follows:
	
	\textit{Step 1:} Because $IG_{n,j}\in d(IG_{j},l_j(k))$ is a sufficient condition for $(1)\in\{(1^{'}),...,(k^{'})\}$ to be satisfied\footnote{In order to {\small$(1)\in\{(1^{'}),...,(k^{'})\}$}, the number of $IG_{n,j}$ which is larger than {\small$IG_{n,(1)}$} is at most $k-1$.}, we use concentration inequalities to derive a lower bound of probability for $IG_{n,j}\in d(IG_{j},l_j(k)),\forall j$, where $d(x,\epsilon)$ denotes the neighborhood of $x$ with radius $\epsilon$.
	
	\textit{Step 2:} By local top-$k$ and global top-$2k$ voting, the most informative attribute $(1)$ will be contained in the global selected set, i.e., $(1)\in\{(1^{'}),...,(k^{'})\}$, if only no less than $[M/2+1]$ local workers select it. We calculate the probability for the case no less than $[M/2+1]$ of all machines select attribute $(1)$ using binomial distribution.
	
	Firstly, we give the probability to ensure {\small$(1)\in\{(1^{'}),...,(k^{'})\}$}.
	We bound the difference between the information gain and the empirical information gain for an arbitrary attribute. To be clear, we will prove, with probability at least {\small$\delta_j(n,k)$}, we have
	{\small\begin{equation*}
		|IG_{n,j}-IG_{j}|\leq l_j(k).
		\end{equation*}}
	For simplify the notations, let $H_j^l(w)=H(Y|w_1\leq X_j\leq w)$, $P_j^l(w)=P(w_1\leq X_j\leq w)$, $H_j^r(w)=H(Y|w\leq X_j\leq w_2)$ and $P_j^r(w)=P(w\leq X_j\leq w_2)$.
	We decompose $\mathcal{H}_{n,j}^l(w_{n,j}^*)-\mathcal{H}_{j}^l(w_{j}^*)$ as
	{\small\begin{eqnarray}
		&\quad&\mathcal{H}_{n,j}^l(w_{n,j}^*)-\mathcal{H}_{j}^l(w_{j}^*)\\
		&=&P_{n,j}^l(w_{n,j}^*)H_{n,j}^l(w_{n,j}^*)-P_j^l(w_j^*)H_j^l(w_j^*)\\
		&=&P_{n,j}^l(w_{n,j}^*)H_{n,j}^l(w_{n,j}^*)-P_{n,j}^l(w_j^*)H_{j}^l(w_j^*)+P_{n,j}^l(w_{j}^*)H_{j}^l(w_{j}^*)-P_j^l(w_j^*)H_j^l(w_j^*). \label{1}
		\end{eqnarray}}
	We decompose $\mathcal{H}_{n,j}^r(w_{n,j^*})-\mathcal{H}_{j}^r(w_{j}^*)$ in a similar way, i.e.,
	{\small\begin{eqnarray}
		&\quad&\mathcal{H}_{n,j}^r(w_{n,j}^*)-\mathcal{H}_{j}^r(w_{j}^*)\\
		&=&P_{n,j}^r(w_{n,j}^*)H_{n,j}^r(w_{n,j}^*)-P_{n,j}^r(w_j^*)H_{j}^l(w_j^*)+P_{n,j}^r(w_{j}^*)H_{j}^r(w_{j}^*)-P_j^r(w_j^*)H_j^r(w_j^*).\label{2}
		\end{eqnarray}}
	By adding Ineq.(\ref{1}) and Ineq.(\ref{2}), we have the following,
	{\small\begin{align*}
		&\quad P(|IG_{n,j}-IG_{j}|>l_j(k))\\
		&= P\left(\left|\mathcal{H}_{n,j}^l(w_{n,j}^*)+\mathcal{H}_{n,j}^r(w_{n,j}^*)-(\mathcal{H}_{j}^l(w_{j}^*)+\mathcal{H}_{j}^r(w_{j}^*))\right|>l_j(k)\right)\\
		&\leq P\left(\left|P_{n,j}^l(w_{j}^*)H_{j}^l(w_{j}^*)-P_j^l(w_j^*)H_j^l(w_j^*)\right|>\frac{l_j(k)}{3}\right)+\\
		&P\left(\left|P_{n,j}^r(w_{j}^*)H_{j}^r(w_{j}^*)-P_j^r(w_j^*)H_j^r(w_j^*)\right|>\frac{l_j(k)}{3}\right)+\\
		&P\left(\left|P_{n,j}^l(w_{n,j}^*)H_{n,j}^l(w_{n,j}^*)-P_{n,j}^l(w_j^*)H_{j}^l(w_j^*)+P_{n,j}^r(w_{n,j}^*)H_{n,j}^r(w_{n,j}^*)-P_{n,j}^r(w_j^*)H_{j}^r(w_j^*)\right|>\frac{l_j(k)}{3}\right)\\
		&\overset{\Delta}{=}I_1+I_2+I_3
		\end{align*}}
	
	For term $I_{1}$, by using Hoeffding's inequality, we have
	{\small\begin{align}
		I_{1}
		&\leq P\left(H_j^l(w_j^*)\times\big|P_j^l(w_j^*)-P_{n,j}^l(w_j^*)\big|>\frac{l_j(k)}{3}\right)\\
		&\leq P\left(\left|P_j^l(w_j^*)-P_{n,j}^l(w_j^*)\right|>\frac{l_j(k)}{3H_j^l(w_j^*)}\right)\\
		&\leq  2\exp{\left(-\frac{2nl_j(k)^2}{9(H_j^l(w_j^*))^2}\right)}
		\end{align}}
	Similarly, for term $I_2$, we have
	{\small\begin{equation}
		I_2\leq 2\exp{\left(-\frac{2nl_j(k)^2}{9(H_j^r(w_j^*))^2}\right)}
		\end{equation}}
	Let $c_j=\min\left\{\frac{2}{9(H_j^l(w_j^*))^2},\frac{2}{9(H_j^l(w_j^*))^2}\right\}$, we have
	{\small\begin{equation}\label{ineq26}
		I_1+I_2\leq 4\exp{\left(-c_jnl_j(k)^2\right)}.
		\end{equation}}
	
	For the term $I_{3}$, 
	we have
	{\small\begin{eqnarray*}
			&&J	\\
			&=&P_{n,j}^l(w_{n,j}^*)H_{n,j}^l(w_{n,j}^*)-P_{n,j}^l(w_j^*)H_{j}^l(w_j^*)+P_{n,j}^r(w_{n,j}^*)H_{n,j}^r(w_{n,j}^*)-P_{n,j}^r(w_j^*)H_{j}^r(w_j^*)\\
			&=&\frac{1}{n}\sum_{i=1}^nI(w_1\leq x_{i,j}\leq w_{n,j}^*)H_{n,j}^l(w_{n,j}^*)+\frac{1}{n}\sum_{i=1}^nI(w_{n,j}^*<x_{i,j}\leq w_2)H_{n,j}^r(w_{n,j}^*)\\
			&&-\frac{1}{n}\sum_{i=1}^nI(w_1\leq x_{i,j}\leq w_{j}^*)H_{j}^l(w_j^*)-\frac{1}{n}\sum_{i=1}^nI(w_{j}^*<x_{i,j}\leq w_2)H_{j}^r(w_{j}^*),\\
		\end{eqnarray*}}
		where $x_{i,j}$ is the $j$-th attribute for the $i$-th instance in the training set.
		
		Let $\Theta$ denote the set of all possible values of $(p_1^l,p_1^r,\cdots,p_{K-1}^l,p_{K-1}^r,w_j)$, where $p_k^l=P(Y=k|w_1\leq X_j\leq w_j)$ and $p_k^r=P(Y=k|w_j<X_j\leq w_2)$. Define the criterion function $\mathbb{M}(\theta)=Pm_{\theta}$, where $m_{\theta}(x,y)=-\log{p_k^l}I(w_1\leq x\leq w_j)-\log{p_k^r}I(w_2\geq x>w_j)$ if $y=k$.  The vector $\theta^*=(p_1^{l*},p_1^{u*},\cdots,p_{K-1}^{l*},p_{K-1}^{u*},w_j^*)$ maximizes $\mathbb{M}(\theta)$, while $\theta_n^*=(p_{n,1}^{l*},p_{n,1}^{r*},\cdots,p_{n,K-1}^{l*},p_{n,K-1}^{r*},w_{n,j}^*)$ minimizes $\mathbb{M}_n(\theta)$. Straightforward algebra shows that
		{\small\begin{eqnarray}
			(m_{\theta}-m_{\theta^*})(X,Y)&=&I(Y=k)[(\log{p_k^{l*}}-\log{p_k^{r*}})(I(w_1\leq X\leq w_{j,n}^*)-I(w_1\leq X<d_j^*))\\
			&&+(\log{p_{n,k}^{l*}}-\log{p_k^{l*}})I(w_1\leq X\leq w_{n,j}^*)\\
			&&+(\log{p_{n,k}^{u*}}-\log{p_k^{r*}})I(w_{n,j}^*\leq X\leq w_2)]
			\end{eqnarray}}
		By following the proof of Theorem 1 in [\cite{banerjee2007confidence}],
		we can get that $n^{2/3}I_3$ converges to $c_2\max_tQ(t)$, where $c_2$ is a constant and $Q(t)$ is composed by the standard two-sided Brownian Motion [\cite{banerjee2007confidence}]. Therefore, we have
		{\small\begin{equation}\label{29}
			P\left(|J|>c_2n^{-\frac{2}{3}}q_{\alpha}\right)<\alpha.
			\end{equation}}
		where  $q_{\alpha}$ is the upper $\alpha$-quantile of  $\max_tQ(t)$.
		Let $c_{2}n^{-\frac{2}{3}}q_{\alpha_j(n)}=\frac{l_j(k)}{3}$. With probability at most $\alpha_j(n)$, we have $IG_{n,j}(w_j^*)-IG_{n,j}>\frac{l_j(k)}{2}$, i.e.,
		{\small\begin{equation}\label{ineq29}
			I_2=P\left(|J|>\frac{l_j(k)}{3}\right)<\alpha_j(n)
			\end{equation}}
		By combining Inequalities (\ref{ineq26}) and (\ref{ineq29}), we have, with probability at most $\delta_j(n,k)=\alpha_j(n)+4\exp{(-c_{j}nl_j(k)^2)}$,
		{\small\begin{equation}\label{28}
			\left|IG_{n,j}-IG_{j}\right|>l_j(k).
			\end{equation}}
		Thus we can get
		{\small\begin{equation}
			P\left(\left|IG_{n,(j)}-IG_{(j)}\right|\leq l_j(k),\forall j\geq k+1\right)\geq 1-\sum_{j=k+1}^{d}{\delta_{(j)}(n,k)}.
			\end{equation}}
		By binomial distribution, we can derive the results in the theorem.\ \ $\Box$	
		
		\vspace{5mm}
		
		\textbf{Proof for regression:}
		
		The proof is similar to classification. We continue to use notations in the previous section and just substitute $IG$ to $VG$.
		
		Similarly, we will prove, with probability at least {\small$\delta_j(n,k)$}, we have
		{\small\begin{equation*}
			|VG_{n,j}-VG_{j}|\leq l_j(k).
			\end{equation*}}
		By the definition of variance gain, we have the following,
		{\small\begin{align*}
			&\quad P\left(|VG_{n,j}-VG_{j}|>l_j(k)\right)\\
			&\leq P(|\sigma_{n,j}^l(w_{n,j}^*)+\sigma_{n,j}^r(w_{n,j}^*)-\sigma_{j}^l(w_j^*)-\sigma_{j}^r(w_j^*)|>l_j(k))\nonumber\\
			&\leq P\left(\left|P_{n,j}^l(w_{j}^*)\sigma_{j}^l(w_{j}^*)-P_j^l(w_j^*)\sigma_j^l(w_j^*)\right|>\frac{l_j(k)}{3}\right)+\\
			&\quad P\left(\left|P_{n,j}^r(w_{j}^*)\sigma_{j}^r(w_{j}^*)-P_j^r(w_j^*)\sigma_j^r(w_j^*)\right|>\frac{l_j(k)}{3}\right)+\\
			&\quad P\left(\left|P_{n,j}^l(w_{n,j}^*)\sigma_{n,j}^l(w_{n,j}^*)-P_{n,j}^l(w_j^*)\sigma_{j}^l(w_j^*)+P_{n,j}^r(w_{n,j}^*)\sigma_{n,j}^r(w_{n,j}^*)-P_{n,j}^r(w_j^*)\sigma_{j}^r(w_j^*)\right|>\frac{l_j(k)}{3}\right)\\
			&\triangleq I_1+I_2+I_3 \\\nonumber
			\end{align*}}
		For term $I_{1}$, by using Hoeffding's inequality, we have
		{\small\begin{align}
			I_{1}
			&\leq P\left(\sigma_j^l(w_j^*)\times\big|P_j^l(w_j^*)-P_{n,j}^l(w_j^*)\big|>\frac{l_j(k)}{3}\right)\nonumber\\
			&\leq P\left(\left|P_j^l(w_j^*)-P_{n,j}^l(w_j^*)\right|>\frac{l_j(k)}{3\sigma_j^l(w_j^*)}\right)\\
			&\leq  2\exp{\left(-\frac{2nl_j(k)^2}{9(\sigma_j^l(w_j^*))^2}\right)}
			\end{align}}
		Similarly, for term $I_2$, we have
		{\small\begin{equation}
			I_2\leq 2\exp{\left(-\frac{2nl_j(k)^2}{9(\sigma_j^r(w_j^*))^2}\right)}
			\end{equation}}
		Let $c_j=\min\left\{\frac{2}{9(\sigma_j^l(w_j^*))^2},\frac{2}{9(\sigma_j^l(w_j^*))^2}\right\}$, we have
		{\small\begin{equation}\label{ineq27}
			I_1+I_2\leq 4\exp{\left(-c_jnl_j(k)^2\right)}.
			\end{equation}}
		
		For the term $I_{3}$, let {\small$J=P_{n,j}^l(w_{n,j}^*)\sigma_{n,j}^l(w_{n,j}^*)-P_{n,j}^l(w_j^*)\sigma_{j}^l(w_j^*)+P_{n,j}^r(w_{n,j}^*)\sigma_{n,j}^r(w_{n,j}^*)-P_{n,j}^r(w_j^*)\sigma_{j}^r(w_j^*)$}.
		According to Theorem 2.2 established by [\cite{banerjee2007confidence}], the following holds,
		{\small\begin{equation}\label{9}
			P\left(|J|>c_2n^{-\frac{2}{3}}q_{\alpha}\right)<\alpha.
			\end{equation}}
		where $c_2$ is a constant for fixed distribution $P$ and $q_{\alpha}$ is the upper $\alpha$-quantile of the standard two-sided Brownian Motion [\cite{banerjee2007confidence}]. With probability at most $\alpha_j(n)$, we have $|J|>\frac{l_j(k)}{3}$, i.e.,
		\begin{equation}\label{ineq9}
		I_3=P\left(|J|>\frac{l_j(k)}{3}\right)<\alpha_j(n)
		\end{equation}
		By combining Ineq.(\ref{ineq27}) and (\ref{ineq9}), we have, with probability at most $\delta_j(n,k)=\alpha_j(n)+4\exp{(-c_{j}nl_j(k)^2)}$,
		{\small\begin{equation}\label{8}
			\left|VG_{n,j}-VG_{j}\right|>l_j(k).
			\end{equation}}
		Thus we can get
		{\small\begin{equation}
			P\left(\left|VG_{n,(j)}-VG_{(j)}\right|\leq h,\forall j\geq k+1\right)\geq 1-\sum_{j=k+1}^d{\delta_{(j)}(n,k)}.
			\end{equation}}
		By binomial distribution, we can derive the results in the theorem.\ \ $\Box$

		\subsection{Theorem 4.2 and its proof}
		
		\textbf{Theorem 4.2:}
		\textit{We denote quantized histogram with $b$ bins of the underlying distribution $P$ as $P^b$, that of the empirical distribution $P_n$ as $P_n^b$, the information gain  of $X_j$ calculated under the distribution $P^b$ and $P_n^b$ as $IG_j^b$ and $IG_{n,j}^b$ respectively, and $f_j(b)\triangleq|IG_j-IG^b_{j}|$. Then, for $\epsilon \leq \min_{j=1,\cdots,d} f_j(b)$, with probability at least $\delta_{j}(n, f_j(b)-\epsilon))$, we have $|IG_{n,j}^b -IG_j|>\epsilon$.}
		
		\textbf{Proof:}\\
		First, $|IG_{n,j}^b-IG_j|=|IG_{n,j}^b-IG^b_j+IG^b_j-IG_j|\geq ||IG_{n,j}^b-IG^b_j|-|f(b)||$. Second, when $n$ is large enough, we have $|f(b)|-|IG_{n,j}^b-IG^b_j|>\epsilon$ with probability $\delta_{j}(n, f_j(b)-\epsilon))$ for $\epsilon \leq \min_{j=1,\cdots,d} f_j(b)$. Thus, the proposition is proven. \ \ $\Box$

\end{document}